\documentclass[runningheads]{llncs}
\usepackage[T1]{fontenc}
\usepackage{graphicx}
\usepackage{amsmath}
\usepackage{amssymb}
\usepackage{booktabs}
\usepackage{bm}
\usepackage{array}
\usepackage{hyperref}
\usepackage{multirow}
\usepackage{orcidlink}
\usepackage{bbding}

\begin{document}
\title{Characterizing Text Branch Sensitivity in Medical Vision-Language Segmentation via Evidence Decoupling}
\titlerunning{Text Sensitivity in Medical VLM Segmentation}
%
\author{Ziquan Liu\orcidlink{0009-0007-3587-1542} \and
Zhewei Zhu\orcidlink{0009-0007-4661-0516} \and
Xuyang Shi\thanks{Corresponding author.}\orcidlink{0000-0002-9473-8692}}

\authorrunning{Z. Liu et al.}

\institute{School of Information and Control Engineering, Southwest University of Science and Technology, Mianyang 621010, China \\
\email{\{ziquanliu, zhuzw\}@mails.swust.edu.cn}, \email{xuyangshi@swust.edu.cn}}

\maketitle

\begin{abstract}
Pretrained vision-language models (VLMs) have shown promising performance in medical image segmentation by incorporating clinical text. However, it remains unclear how much textual information actually contributes to pixel-level predictions. In this work, we systematically investigate the role of text in multimodal medical image segmentation. We first analyze several commonly used fusion strategies and find that segmentation performance is largely insensitive to the choice of fusion module. To further understand modality interactions, we propose an Evidence Decoupling Decoder (EDD) based on evidential deep learning and deep supervision. EDD serves as an internal representation analysis tool that decomposes image evidence and text-modulated evidence throughout the decoding process while maintaining competitive segmentation performance. Experimental results show that the sensitivity to text perturbation varies substantially across datasets. On BUSI and BTMRI, removing text causes catastrophic performance drops, indicating strong model reliance on textual input. On ISIC and Kvasir-SEG, text exerts relatively marginal influence. We further find that text affects predictions mainly through global semantic modulation rather than independent spatial localization, and that the specific semantic components driving text sensitivity differ across datasets. These findings provide a deeper understanding of modality interaction in multimodal medical image segmentation and offer practical insights for future model design.

\end{abstract}

\section{Introduction}

Pretrained vision-language models (VLMs) have advanced multimodal medical image segmentation by incorporating clinical text into vision networks~\cite{li2023lvit,koleilat2026medclipseg}. However, existing work implicitly assumes text is a beneficial auxiliary modality and focuses on designing increasingly complex fusion architectures~\cite{yang2022lavt,liu2023sat,tomar2022tganet,liu2024mlip}. While recent studies have identified modality imbalance and collapse as structural properties of VLM training~\cite{chaudhuri2025collapse,kwon2025seesaw,yaras2024modalitygap}, these operate at the representation or gradient level and do not characterize per-pixel modality interaction in medical segmentation. A more basic question remains understudied: how much does the text branch actually contribute to pixel-level decisions?


To study this, we evaluate multimodal medical segmentation across four imaging modalities~\cite{aldhabyani2020busi,cheng2017btmri,jha2020kvasirseg,codella2018isic2017,codella2019isic2018}, five pretrained medical VLMs~\cite{radford2021learning,wang2022medclip,zhang2023biomedclip,eslami2023pubmedclip,khattak2024unimedclip} and four cross-modal fusion operators~\cite{perez2018film,vaswani2017attention,hu2018senet,ronneberger2015unet}. Our experiments reveal that four topologically distinct fusion operators yield very similar segmentation results, suggesting that changing the fusion micro-structure has little impact on multimodal information use, as visual features dominate under current pretrained representations.

To examine internal modality interactions, we introduce evidential deep learning (EDL)~\cite{sensoy2018edl} and propose a lightweight Evidence Decoupling Decoder (EDD). EDL reformulates deterministic probability outputs as non-negative evidence under a Dirichlet distribution, allowing us to characterize the learned modality representations. Inspired by deep supervision~\cite{lee2015deeplysupervised}, we design separate vision and text evidence heads at each upsampling layer of the decoder. These heads extract image evidence $\mathbf{e}_I$ from visual-only features and text-modulated evidence $\mathbf{e}_T$, driven by a unified EDL joint loss. Experiments confirm that this decoder preserves segmentation accuracy close to the standard decoder across all VLM and dataset combinations, providing an interpretability tool that reveals how the model internally represents modality interaction.
In summary, this paper makes three contributions:

\begin{itemize}
\item We show that four topologically distinct fusion operators yield nearly identical segmentation performance across five VLMs and four medical imaging datasets, indicating that fusion design has limited impact under current pretrained representations.
\item We propose an Evidence Decoupling Decoder (EDD) that combines evidential deep learning with deep supervision, achieving per-layer decomposition of image and text-modulated evidence without sacrificing accuracy.
\item We reveal through perturbation analysis that text sensitivity is strongly dataset dependent. On BTMRI and BUSI, text corruption causes severe degradation, while the impact of text is relatively small on ISIC and Kvasir-SEG. Text-modulated evidence is globally distributed, and different text components drive sensitivity across datasets.
\end{itemize}
\section{Related Work}

\subsection{Text-Guided Medical Image Segmentation}
Vision-language models (VLMs) increasingly leverage clinical text as auxiliary guidance for medical image segmentation. Following foundational architectures such as ViLT~\cite{kim2021vilt}, text-guided segmentation frameworks typically adopt either early fusion of linguistic and visual features within the encoder~\cite{yang2022lavt,li2023lvit}, or text-conditioned attention and contrastive alignment in the decoder~\cite{tomar2022tganet,liu2023sat,liu2024mlip}. MedCLIPSeg~\cite{koleilat2026medclipseg} further introduces probabilistic adaptation for data-efficient multimodal segmentation. While these methods advance architecture design, they uniformly treat text as an inherently beneficial modality, without quantitatively investigating how much the text branch actually contributes to pixel-level decisions.

\subsection{Modality Interaction and Imbalance in Vision-Language Models}
Recent work questions whether multimodal learning always yields balanced contributions. Studies characterize the modality gap in contrastive VLMs, where image and text embeddings form distinct clusters in the joint space, and link it to mismatched training pairs, temperature, and information imbalance between modalities~\cite{yaras2024modalitygap,schrodi2024twoeffects}. At the fusion level, modality collapse has been attributed to noisy cross‑modal feature entanglement and unaligned inter‑modal gradients that prevent balanced convergence~\cite{chaudhuri2025collapse,kwon2025seesaw}. In medical segmentation, TMCA~\cite{li2024tmca} uses contrastive alignment to narrow image‑text pattern gaps. While these works establish modality dominance as a structural property of VLM training, they operate at the representation or gradient level and do not offer per‑pixel decomposition of modality‑specific spatial contributions.

\subsection{Evidential Deep Learning for Medical Image Analysis}
Evidential deep learning (EDL)~\cite{sensoy2018edl} casts classification as evidence under a Dirichlet distribution, enabling joint prediction and uncertainty quantification. Recent EDL-based medical segmentation methods span semi-supervised learning~\cite{he2024evidentialfusion}, brain tumor segmentation~\cite{li2022edlbraintumor}, and progressive uncertainty guidance~\cite{yang2025edlkan}. Building on Dempster–Shafer theory, Fidon et al.~\cite{fidon2024dempstershafer} and Huang et al.~\cite{huang2023deepevidentialfusion} fuse imaging modalities (e.g., PET/CT) with reliability weights, while DEviS~\cite{zou2025devis} and SURE~\cite{li2025sure} improve uncertainty interpretability and calibration. However, existing multimodal EDL frameworks focus on multiple imaging modalities rather than vision-language pairs, and none decompose decisions into per-modality, per-layer evidence streams. Our evidence decoupling decoder fills this gap.
\section{Methodology}

\subsection{Problem Definition and Overall Framework}

\begin{figure*}[ht]
\centering
\includegraphics[width=0.95\linewidth]{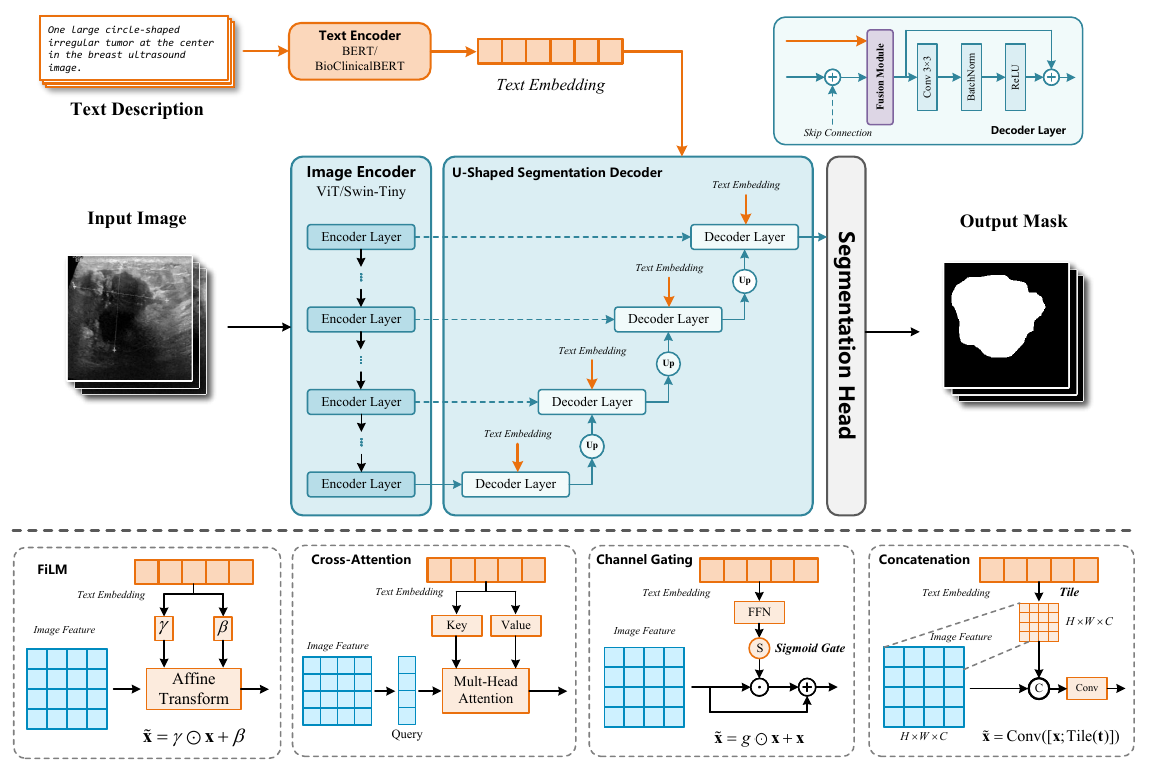}
\caption{Our baseline framework feeds image samples and corresponding text into the image encoder and text encoder, respectively. The resulting features are then processed step by step through a U-shaped segmentation decoder for segmentation. The lower half of the figure provides a schematic of the four fusion operators.}
\label{fig:method}
\end{figure*}

Given an input image $\mathbf{I} \in \mathbb{R}^{H \times W \times 3}$ and its text description $\mathbf{T}$, we aim to predict a binary segmentation mask $\hat{\mathbf{Y}} \in \{0, 1\}^{H \times W}$. As shown in Fig.~\ref{fig:method}, the overall framework consists of an image encoder $f_I$, a text encoder $f_T$, and a segmentation decoder $D$:
\begin{equation}
\hat{\mathbf{Y}} = D\left(f_I(\mathbf{I}),\; f_T(\mathbf{T})\right) \label{eq:framework}
\end{equation}
To enable fine-grained analysis of modality-specific representations, we propose an Evidence Decoupling Decoder (EDD) that separately models image and text evidence streams throughout the decoding process.

\subsection{Baseline Segmentation Network with Multimodal Features}

To cover different vision-language pretraining paradigms, we evaluate five representative pretrained medical VLMs as image-text feature extractors. Their configurations are listed in Table~\ref{tab:encoders}.

\begin{table}[t]
\centering
\caption{Architecture and pretraining details of evaluated encoders.}
\label{tab:encoders}
\resizebox{1.0\linewidth}{!}{
\begin{tabular}{l | l l l l l}
\toprule
\textbf{Model} & \textbf{Vision Backbone} & \textbf{Text Backbone} & \textbf{Pretraining Data} & \textbf{Scale} & \textbf{Modality Coverage} \\
\midrule
CLIP~\cite{radford2021learning} & ViT-B/16 & Transformer & WIT & $\sim$400M & Natural images \\
MedCLIP~\cite{wang2022medclip} & Swin-Tiny & BioClinicalBERT & MIMIC-CXR & $\sim$200K & Chest X-ray only \\
BioMedCLIP~\cite{zhang2023biomedclip} & ViT-B/16 & PubMedBERT & PMC-15M & $\sim$15M & Biomedical images \\
PubMedCLIP~\cite{eslami2023pubmedclip} & ViT-B/32 & Transformer & PubMed+ROCO & 300K--400K & Multimodal medical \\
UniMed-CLIP~\cite{khattak2024unimedclip} & ViT-B/16 & BiomedBERT & Mixed medical corpora & $\sim$5.3M & All medical modalities \\
\bottomrule
\end{tabular}
}
\end{table}

The standard decoder uses a U-shaped hierarchical structure that gradually recovers spatial resolution through skip connections. Let $l$ index decoder levels from the shallowest ($l = 0$) to the deepest ($l = L - 1$). At level $l$, we upsample the deeper-level feature and fuse it with the encoder skip feature $\mathbf{F}^{(l)}$ to obtain an intermediate visual feature $\mathbf{u}^{(l)}$:
\begin{equation}
\mathbf{u}^{(l)} = \text{Up}\left(\mathbf{x}^{(l+1)}\right) + \mathbf{F}^{(l)}, \label{eq:skip}
\end{equation}
where $\text{Up}(\cdot)$ is the upsampling operation. After fusing image features with skip connections, the fusion module injects text embedding $\mathbf{t}\in \mathbb{R}^{D}$ into the visual features $\mathbf{u}^{(l)}$. A residual convolutional block refines the result to produce the current level output $\mathbf{x}^{(l)}$. The shallowest layer output $\mathbf{x}^{(0)}$ passes through a simple segmentation head to produce the segmentation result.

To study the effect of different cross-modal interaction mechanisms, we integrate four typical fusion operators into the standard decoder~\cite{perez2018film,vaswani2017attention,hu2018senet,ronneberger2015unet}, as shown in the lower part of Fig.~\ref{fig:method}.

\subsection{Evidence Decoupling Decoder}

\subsubsection{Design Motivation and Modeling Considerations.} A standard softmax-based decoder outputs deterministic class probabilities and cannot separate modality-specific representations. We adopt the EDL framework~\cite{sensoy2018edl}, which models outputs as non-negative evidence under a Dirichlet distribution, enabling explicit decomposition of modality-level evidence. Inspired by deep supervision~\cite{lee2015deeplysupervised}, we aim to capture the per-layer evolution of evidence from both branches. The evidence decoupling decoder is guided by two design principles:

\begin{description}
\item[Evidence-space projectability.] Intermediate features at each decoder level are semantically rich enough to be projected into non-negative evidence via a lightweight convolution and Softplus activation.
\item[Hierarchical supervision consistency.] Per-level evidence maps share the same semantic target as the final output, differing only in spatial resolution, so all levels can use the same supervision signal.
\end{description}

Guided by these principles, we design an evidence decoupling branch, dividing the evidence at each layer into pure image evidence, text-modulated evidence, and fused evidence.

\subsubsection{Evidence Decoupling Branch Design.} To avoid contaminating raw visual representations, the evidence decoupling branch runs after each skip connection but before the fusion module, where the visual feature $\mathbf{u}^{(l)}$ is not yet text-modulated and thus provides a suitable base for evidence extraction.

\paragraph{Image evidence stream.} The image evidence head consists of a $1 \times 1$ convolution and a Softplus activation. It directly extracts a non-negative image evidence map $\mathbf{e}_I^{(l)} \in \mathbb{R}_{+}^{K \times H_l \times W_l}$ from the raw visual feature, where $K = 2$ is the number of classes and $H_l, W_l$ denote the spatial resolution at decoder level $l$:
\begin{equation}
\mathbf{e}_I^{(l)} = \text{Softplus}\left(\text{Conv}_{1 \times 1}\left(\mathbf{u}^{(l)}\right)\right) \label{eq:eI}
\end{equation}
This stream depends only on visual features. It reflects how well the vision encoder can independently discriminate the segmentation target.

\paragraph{Text evidence stream.} Unlike image features, the text embedding $\mathbf{t}\in \mathbb{R}^{D}$ is a global semantic vector. It lacks spatial structure and per-pixel correspondence, so it cannot directly produce a spatially resolved evidence map. To bridge this gap, we model text evidence as a global channel-level modulation on image spatial features, where the text provides class-wise semantic intensity coefficients and the image provides spatial response bases. Their product spatializes the text semantics.

First, an MLP maps the text embedding $\mathbf{t}\in \mathbb{R}^{D}$ into $K$ dimension semantic intensity coefficients:
\begin{equation}
\mathbf{e}_T^{\text{chan}} = \text{Softplus}\left(\text{MLP}(\mathbf{t})\right) \in \mathbb{R}_{+}^{K} \label{eq:eTchan}
\end{equation}
At the same time, we extract spatial evidence bases $\mathbf{s}^{(l)} \in \mathbb{R}_{+}^{K \times H_l \times W_l}$ from the current-level visual feature:
\begin{equation}
\mathbf{s}^{(l)} = \text{Softplus}\left(\text{Conv}_{1 \times 1}\left(\mathbf{u}^{(l)}\right)\right) \label{eq:s}
\end{equation}

Note that although $\mathbf{s}^{(l)}$ and $\mathbf{e}_I^{(l)}$ both use a $1 \times 1$ convolution plus Softplus on the same input $\mathbf{u}^{(l)}$, they have independent parameters and distinct semantic roles. $\mathbf{e}_I^{(l)}$ learns to provide direct evidence for the segmentation target from the image modality. $\mathbf{s}^{(l)}$ learns spatial response bases that are suitable for modulation by text semantics.

The text-modulated evidence $\mathbf{e}_T^{(l)}$ is then computed by channel-wise broadcast multiplication:
\begin{equation}
\mathbf{e}_T^{(l)} = \mathbf{e}_T^{\text{chan}} \odot \mathbf{s}^{(l)} \in \mathbb{R}_+^{K\times H_l \times W_l} \label{eq:eT}
\end{equation}
This design lets text evidence inherit visual spatial patterns in structure while being globally regulated by text semantic coefficients in intensity.

It is important to clarify that $\mathbf{e}_T^{(l)}$ does not represent a purely text-driven evidence map that would exist independently of vision. Text evidence is, by construction, text-modulated visual evidence. Its spatial structure is inherited from $\mathbf{s}^{(l)}$, which derives from visual features $\mathbf{u}^{(l)}$. The decomposition primarily characterizes how the trained decoder represents modality interaction, rather than measuring an intrinsic modality-independent text contribution. This architectural choice is intentional. A global text vector has no native spatial resolution, so spatialization through visual features is necessary. However, this choice imposes an interpretive constraint. The resulting evidence maps reflect the model's learned encoding of text semantics in visual coordinates, not an independent text-to-pixel pathway.

\paragraph{Evidence fusion.} The two decoupled evidence streams are combined additively to form the total fused evidence $\mathbf{e}_{\text{add}}^{(l)}$ at the current level:
\begin{equation}
\mathbf{e}_{\text{add}}^{(l)} = \mathbf{e}_I^{(l)} + \mathbf{e}_T^{(l)} \label{eq:eadd}
\end{equation}
This additive combination requires no learned parameters, making the decomposition structurally transparent by construction. We emphasize that this decomposition reflects the model's learned evidence representation rather than ground-truth modality attribution.

\subsection{Loss Function and Inference Quantification}

\subsubsection{Baseline Network Loss.} For the standard segmentation network without EDL branches, we use the combination of Dice loss and cross-entropy (CE) loss:
\begin{equation}
\mathcal{L}_{\text{seg}} = \mathcal{L}_{\text{Dice}} + \lambda \mathcal{L}_{\text{CE}}, \label{eq:segLoss}
\end{equation}
where $\lambda$ is the loss balance parameter.

\subsubsection{Evidence Network Joint Loss.} With EDL, segmentation is reformulated as an evidence regression process under the Dirichlet distribution. Given an evidence vector $\mathbf{e}$, the corresponding Dirichlet concentration parameters are $\boldsymbol{\alpha} = \mathbf{e} + 1$, and the total evidence is $S = \sum_{k=1}^{K} \alpha_k$. The single-layer EDL loss $\mathcal{L}_{\text{EDL}}(\mathbf{e}, Y)$ consists of a fidelity term and a KL regularization term.

\paragraph{Fidelity term.} For the true class $y$, we minimize the expected squared error under the Dirichlet distribution:
\begin{equation}
\mathcal{L}_{\text{err}}(\mathbf{e}, Y) = \sum_{k=1}^{K} \left( y_k - \frac{\alpha_k}{S} \right)^2 + \frac{\alpha_k (S - \alpha_k)}{S^2 (S + 1)} \label{eq:Lerr}
\end{equation}
The first term measures the deviation between the expected probability and the ground-truth label. The second term accounts for prediction variance.

\paragraph{KL regularization term.} For non-true classes, we use KL divergence to push their evidence toward zero. This prevents the model from assigning high confidence to all classes:
\begin{equation}
\mathcal{L}_{\text{KL}}(\mathbf{e}, Y) = \text{KL}\left[\text{Dir}(\mathbf{p} \mid \tilde{\boldsymbol{\alpha}})\;\|\; \text{Dir}(\mathbf{p} \mid \mathbf{1})\right], \label{eq:LKL}
\end{equation}
where $\tilde{\boldsymbol{\alpha}} = Y + (1 - Y) \odot \boldsymbol{\alpha}$ is the Dirichlet parameter after removing the evidence of the true class.

The single-layer EDL loss is then:
\begin{equation}
\mathcal{L}_{\text{EDL}}(\mathbf{e}, Y) = \mathcal{L}_{\text{err}}(\mathbf{e}, Y) + \lambda_t \cdot \mathcal{L}_{\text{KL}}(\mathbf{e}, Y), \label{eq:LEDL}
\end{equation}
where $\lambda_t$ is an annealing coefficient that gradually increases during training to avoid excessive regularization early on.

\paragraph{Multi-level joint loss.} The overall training loss is a weighted sum of the EDL loss at the final segmentation head and at all intermediate levels ($L = 4$ decoder levels):
\begin{equation}
\mathcal{L} = \mathcal{L}_{\text{EDL}}\left(\mathbf{e}^{\text{final}}, Y\right) + \sum_{l=0}^{L-1} w_l \cdot \mathcal{L}_{\text{EDL}}\left(\mathbf{e}_{\text{add}}^{(l)}, Y\right) \label{eq:Ltotal}
\end{equation}
where the weights $w_l$ decrease with increasing layer depth.

At inference time, the model outputs expected class probabilities $\hat{p}_k = \alpha_k / S$. Additionally, the epistemic uncertainty at each pixel can be quantified from the total evidence:
\begin{equation}
u = \frac{K}{S} = \frac{K}{\sum_{j=1}^{K} \alpha_j} \label{eq:uncertainty}
\end{equation}
When a pixel has low total evidence $S$, the model lacks sufficient support for its decision, corresponding to high epistemic uncertainty.

\section{Experiments}
\label{sec:experiments}

\subsection{Experimental Setup}
\label{sec:exp_setup}

We evaluate on four medical imaging datasets with paired text descriptions, including BUSI~\cite{aldhabyani2020busi}, BTMRI~\cite{cheng2017btmri}, Kvasir-SEG~\cite{jha2020kvasirseg}, and ISIC~\cite{codella2018isic2017,codella2019isic2018}. The image-text pairs for all four datasets are obtained from the MedCLIPSeg~\cite{koleilat2026medclipseg} repository, which provides curated text descriptions aligned with each sample. To ensure fair comparison, the dataset splits also follow those provided by MedCLIPSeg, and all experiments adopt the same training strategy and optimization hyperparameters. Both the image encoder and text encoder are not frozen throughout training. All experiments use images resized to $224\times 224$, batch size 32, AdamW optimizer with learning rate $2\times 10^{-4}$ and weight decay $1\times 10^{-4}$, trained and tested on an NVIDIA V100 32G GPU. Segmentation performance is measured by Dice coefficient and IoU.

\subsection{Baseline Performance Analysis}
\label{sec:baseline}

\subsubsection{Impact of Different VLM Encoders.} We first evaluate the baseline segmentation performance of five pretrained VLM encoders across four datasets, as shown in Table~\ref{tab:vlm_benchmark}. The Gating fusion operator is used as the default in all experiments in this section.

Substantial performance gaps exist among VLM encoders. For example, on BTMRI, MedCLIP achieves 91.61\% Dice, while PubMedCLIP reaches only 78.40\% and BioMedCLIP attains 87.63\%. The best-performing encoder also depends on the dataset, as UniMed-CLIP leads on BUSI and ISIC while MedCLIP excels on BTMRI and Kvasir-SEG. No single VLM encoder consistently dominates all datasets.

\begin{table}[t]
\centering
\caption{Segmentation performance of different VLM encoders across four datasets. Best results are in bold.}
\label{tab:vlm_benchmark}
\resizebox{\linewidth}{!}{
\begin{tabular}{l|cc|cc|cc|cc}
\toprule
& \multicolumn{2}{c|}{BTMRI} & \multicolumn{2}{c|}{BUSI} & \multicolumn{2}{c|}{Kvasir-SEG} & \multicolumn{2}{c}{ISIC} \\
\cmidrule(lr){2-3} \cmidrule(lr){4-5} \cmidrule(lr){6-7} \cmidrule(lr){8-9}
\textbf{Model} & Dice & IoU & Dice & IoU & Dice & IoU & Dice & IoU \\
\midrule
UNet~\cite{ronneberger2015unet} & 86.34 & 79.88 & 68.12 & 60.73 & 80.87 & 71.64 & 88.39 & 82.41 \\
UNet++~\cite{zhou2018unetpp} & 85.76 & 77.71 & 67.22 & 59.48 & 81.75 & 73.30 & 88.47 & 82.84 \\
DeepLabv3~\cite{chen2017deeplabv3} & 85.25 & 78.63 & 65.81 & 58.23 & 82.03 & 72.58 & 81.94 & 70.79 \\
TransUNet~\cite{chen2021transunet} & 71.47 & 60.36 & 62.80 & 54.74 & 60.26 & 50.08 & 85.18 & 77.63 \\
\midrule
CLIP~\cite{radford2021learning} & 87.85 & 80.07 & 81.22 & 68.88 & 85.60 & 74.85 & 92.37 & 85.85 \\
MedCLIP~\cite{wang2022medclip} & \textbf{91.61} & \textbf{85.42} & 82.43 & 70.68 & \textbf{91.32} & \textbf{84.06} & \textbf{93.71} & \textbf{88.18} \\
PubMedCLIP~\cite{eslami2023pubmedclip} & 78.40 & 68.46 & 81.12 & 68.58 & 76.93 & 62.53 & 91.35 & 84.11 \\
BioMedCLIP~\cite{zhang2023biomedclip} & 87.63 & 79.69 & \textbf{83.93} & \textbf{72.89} & 88.74 & 79.81 & 92.77 & 86.55 \\
UniMed-CLIP~\cite{khattak2024unimedclip} & 88.84 & 81.32 & 82.99 & 71.45 & 89.32 & 80.73 & 93.60 & 87.98 \\
\bottomrule
\end{tabular}
}
\end{table}

\subsubsection{Fusion Module Sensitivity Analysis.} To systematically evaluate the impact of cross-modal fusion mechanisms, we compare four fusion operators (FiLM~\cite{perez2018film}, cross-attention~\cite{vaswani2017attention}, channel gating~\cite{hu2018senet}, concatenation~\cite{ronneberger2015unet}) across all VLM encoders.

\begin{table}[t]
\centering
\caption{Average Dice scores (\%) across four datasets for different fusion modules, along with the maximum fluctuation range $\Delta_{\max}$ (\%).}
\label{tab:fusion_sensitivity}
\resizebox{\linewidth}{!}{
\begin{tabular}{l|cccc|c}
\toprule
\textbf{Model} & \textbf{FiLM} & \textbf{Cross-Attention} & \textbf{Gating} & \textbf{Concat} & $\mathbf{\Delta_{\max}}$ \\
\midrule
CLIP & 85.45 & 86.01 & 86.76 & 85.57 & 1.31 \\
MedCLIP & 89.81 & 89.56 & 89.77 & 89.29 & 0.52 \\
PubMedCLIP & 82.02 & 82.26 & 81.95 & 82.47 & 0.52 \\
BioMedCLIP & 86.42 & 85.52 & 88.27 & 85.53 & 2.75 \\
UniMed-CLIP & 88.54 & 88.91 & 88.69 & 89.26 & 0.72 \\
\bottomrule
\end{tabular}
}
\end{table}

As shown in Table~\ref{tab:fusion_sensitivity}, fusion modules cause only marginal performance variation, with the maximum fluctuation $\Delta_{\max}$ of 2.75\% for BioMedCLIP. The gaps between different VLMs, such as the roughly 4\% difference between CLIP and MedCLIP on BTMRI, far exceed the differences across fusion strategies. This suggests that segmentation quality depends primarily on visual feature quality rather than fusion topology, and that merely optimizing the fusion module cannot fundamentally alter multimodal information utilization.

\subsection{Evidence Decoupling Decoder Validation}
\label{sec:edl_validation}

\begin{table}[t]
\centering
\caption{Segmentation performance comparison between the standard VLM decoder and the evidence decoupling decoder (Dice / IoU, \%).}
\label{tab:edl_decoder}
\resizebox{\linewidth}{!}{
\begin{tabular}{l|cc|cc|cc|cc}
\toprule
& \multicolumn{2}{c|}{BTMRI} & \multicolumn{2}{c|}{BUSI} & \multicolumn{2}{c|}{Kvasir-SEG} & \multicolumn{2}{c}{ISIC} \\
\cmidrule(lr){2-3} \cmidrule(lr){4-5} \cmidrule(lr){6-7} \cmidrule(lr){8-9}
\textbf{Model} & Dice & IoU & Dice & IoU & Dice & IoU & Dice & IoU \\
\midrule
CLIP & 87.85 & 80.07 & 81.22 & 68.88 & 85.60 & 74.85 & 92.37 & 85.85 \\
CLIP + EDD & 87.50 & 79.52 & 83.30 & 72.05 & 86.46 & 76.22 & 92.17 & 85.50 \\
\midrule
MedCLIP & 91.61 & 85.42 & 82.43 & 70.68 & 91.32 & 84.06 & 93.71 & 88.18 \\
MedCLIP + EDD & 91.61 & 85.41 & 84.34 & 78.07 & 90.46 & 82.61 & 93.12 & 87.17 \\
\midrule
PubMedCLIP & 78.40 & 68.46 & 81.12 & 68.58 & 76.93 & 62.53 & 91.35 & 84.11 \\
PubMedCLIP + EDD & 84.72 & 75.83 & 79.82 & 67.16 & 82.06 & 69.60 & 92.05 & 85.29 \\
\midrule
BioMedCLIP & 87.63 & 79.69 & 83.93 & 72.89 & 88.74 & 79.81 & 92.77 & 86.55 \\
BioMedCLIP + EDD & 88.55 & 81.01 & 81.78 & 69.62 & 87.80 & 78.26 & 92.55 & 86.19 \\
\midrule
UniMed-CLIP & 88.84 & 81.32 & 82.99 & 71.45 & 89.32 & 80.73 & 93.60 & 87.98 \\
UniMed-CLIP + EDD & 89.46 & 82.24 & 81.17 & 68.91 & 90.44 & 82.60 & 93.39 & 87.62 \\
\bottomrule
\end{tabular}
}
\end{table}

We verify that the proposed Evidence Decoupling Decoder (EDD) preserves segmentation performance while enabling evidence decomposition. Table~\ref{tab:edl_decoder} compares the standard VLM decoder against EDD. For nearly all VLM-dataset pairs, the EDD segmentation performance remains close to that of the standard decoder, with Dice and IoU deviations generally modest, although a few cases exhibit larger changes. In most cases, the EDD variant yields gains over the standard decoder, for example a 2.08\% Dice improvement for CLIP on BUSI and a 6.32\% gain for PubMedCLIP on BTMRI. A few pairs show slight degradations, such as BioMedCLIP on BUSI dropping by 2.15\% and UniMed-CLIP on BUSI by 1.82\%, which are consistent with the added EDL regularization and remain within the expected variance for medical image segmentation. The occasional performance improvements may be due to the effect of the auxiliary deep supervision signal, which can improve and optimize the baseline decoder when it is underfitting. EDD's primary purpose remains evidence decomposition for interpretability rather than performance enhancement. Overall, these results demonstrate that the evidence decoupling branch can be introduced without systematically disrupting the original segmentation optimization.

\subsection{Evidence Decoupling Visualization Analysis}
\label{sec:edl_viz}

\begin{figure*}[t]
\centering
\includegraphics[width=0.95\linewidth]{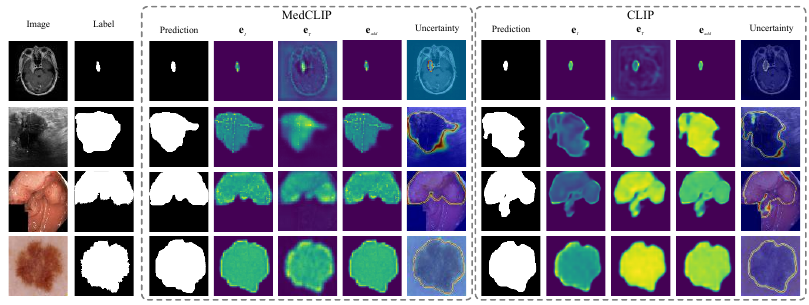}
\caption{Decoupled evidence maps from the last decoder layer of CLIP and MedCLIP.}
\label{fig:evidence_last_layer}
\end{figure*}

Fig.~\ref{fig:evidence_last_layer} shows the evidence maps $\mathbf{e}_I$, $\mathbf{e}_T$, and $\mathbf{e}_{\text{add}}$ from the final decoder layer of CLIP and MedCLIP. Image evidence $\mathbf{e}_I$ concentrates on lesion regions and provides precise spatial localization. Text evidence $\mathbf{e}_T$ generates no independent spatial responses but is superimposed on image features as a smooth global modulator, so it cannot autonomously correct spatial deviations in image features. Notably, text evidence can activate textures missing from image evidence. On the BTMRI dataset, MedCLIP's $\mathbf{e}_T$ clearly responds to fine-grained structures that $\mathbf{e}_I$ does not capture, which indicates that text semantics can complement local patterns missing from the visual representation. The additive evidence $\mathbf{e}_{\text{add}}$ remains nearly identical to $\mathbf{e}_I$, confirming that image evidence dominates the fused output. Comparing the two models, MedCLIP's Swin-Tiny backbone yields finer-grained evidence, while for CLIP, text evidence visibly enhances weak visual responses. These visualizations illustrate EDD as a qualitative tool that reveals how image and text interact differently across VLM encoders and datasets.

\begin{table}[ht]
\centering
\caption{Impact of text perturbation on MedCLIP and CLIP segmentation performance (Dice / IoU, \%).}
\label{tab:text_perturb}
\resizebox{\linewidth}{!}{
\begin{tabular}{ll|cc|cc|cc|cc}
\toprule
& & \multicolumn{2}{c|}{BTMRI} & \multicolumn{2}{c|}{BUSI} & \multicolumn{2}{c|}{Kvasir-SEG} & \multicolumn{2}{c}{ISIC} \\
\cmidrule(lr){3-4} \cmidrule(lr){5-6} \cmidrule(lr){7-8} \cmidrule(lr){9-10}
\textbf{Model} & \textbf{Text Condition} & Dice & IoU & Dice & IoU & Dice & IoU & Dice & IoU \\
\midrule
\multirow{5}{*}{MedCLIP} & Original & 91.61 & 85.42 & 82.43 & 70.68 & 91.32 & 84.06 & 93.71 & 88.18 \\
& w/o Text & 40.76 & 40.58 & 54.50 & 46.68 & 78.86 & 69.47 & 89.76 & 80.16 \\
& Random & 60.05 & 54.95 & 56.06 & 47.71 & 88.93 & 82.37 & 91.81 & 85.52 \\
& Missing Location & 90.88 & 85.62 & 71.58 & 63.47 & 89.60 & 83.56 & 92.51 & 86.63 \\
& Contradictory & 14.73 & 12.26 & 59.71 & 51.06 & 88.68 & 82.08 & 91.87 & 81.93 \\
\midrule
\multirow{5}{*}{CLIP} & Original & 87.85 & 80.07 & 81.22 & 68.88 & 85.60 & 74.85 & 92.37 & 85.85 \\
& w/o Text & 54.25 & 50.69 & 41.60 & 33.83 & 64.78 & 54.65 & 49.37 & 39.32 \\
& Random & 59.09 & 54.90 & 42.77 & 35.08 & 69.52 & 59.78 & 85.43 & 77.01 \\
& Missing Location & 82.84 & 76.21 & 56.10 & 47.87 & 75.63 & 66.08 & 90.64 & 83.89 \\
& Contradictory & 23.09 & 15.01 & 49.01 & 39.79 & 58.76 & 46.89 & 82.08 & 72.61 \\
\bottomrule
\end{tabular}
}
\end{table}

\subsection{Text Perturbation Experiments}
\label{sec:text_perturb}

Visual analysis alone is insufficient to quantify text branch impact on segmentation performance. This section designs a text perturbation experiment using five text conditions during inference, applied to already-trained multimodal segmentation models.

\begin{enumerate}
\item \textbf{Original Text} refers to the original, correct text description.
\item \textbf{w/o Text} indicates complete removal of text input.
\item \textbf{Random Text} means a randomly matched, unrelated text description from another sample.
\item \textbf{Missing Location} removes lesion location information from the text.
\item \textbf{Contradictory Text} supplies internally inconsistent descriptions that can confuse the model.
\end{enumerate}

These text perturbation conditions follow the design of MedCLIPSeg~\cite{koleilat2026medclipseg}, where the Missing Location condition removes spatial cues (e.g., lesion position descriptors) from the prompt, and the Contradictory condition supplies internally inconsistent descriptions.

\begin{figure*}[ht]
\centering
\includegraphics[width=0.95\linewidth]{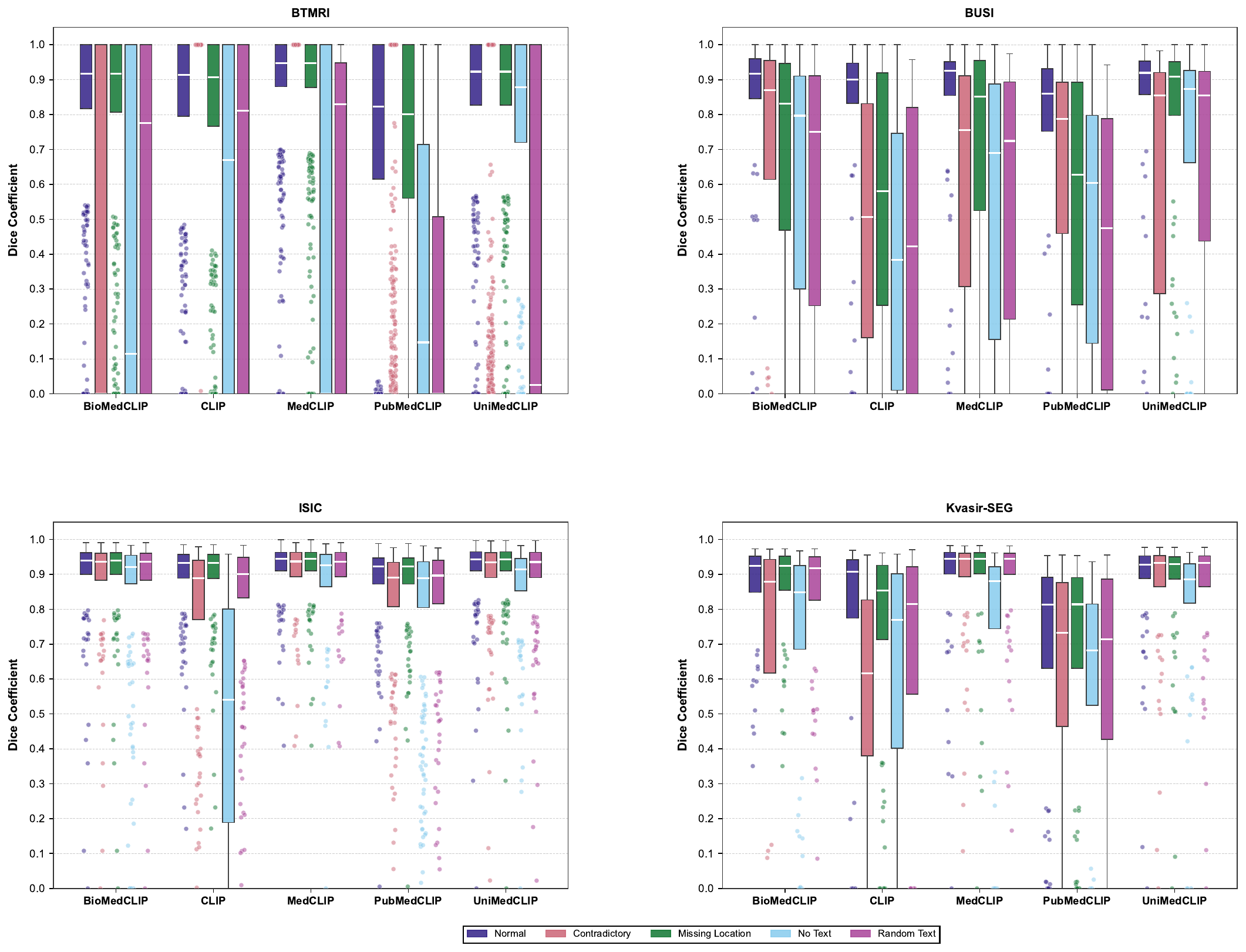}
\caption{Box plots of the Dice metric for five models tested on four datasets with text perturbations.}
\label{fig:dice_box}
\end{figure*}

As observed in Table~\ref{tab:text_perturb}, removing text entirely leads to a drastic drop in performance. For example, MedCLIP on BTMRI falls from 91.61\% to 40.76\% Dice, confirming that text supplies complementary constraints beyond those available from visual features alone. Conversely, supplying contradictory text causes further degradation, with MedCLIP's Dice on BTMRI falling to 14.73\%. However, the sensitivity to text perturbation is strongly dataset-dependent. ISIC demonstrates marked robustness, with MedCLIP retaining 89.76\% Dice even without text. In contrast, BTMRI and BUSI exhibit much higher sensitivity to the quality of the accompanying text. These findings support a differentiated view of text sensitivity rather than a uniform text contribution. On BTMRI and BUSI, text perturbation causes severe performance degradation, indicating strong model reliance on textual input. On ISIC, MedCLIP remains robust without text (89.76\% Dice), whereas CLIP experiences a large drop (49.37\%), demonstrating that text sensitivity is also model-dependent. On Kvasir-SEG, both models show moderate degradation, with MedCLIP falling to 78.86\% and CLIP to 64.78\%.

Furthermore, the perturbation patterns reveal that different text components play distinct roles across datasets. On BTMRI, removing location descriptors barely affects MedCLIP performance, with Dice moving from 91.61\% to 90.88\%, yet removing all text causes a catastrophic drop to 40.76\%. This suggests that text sensitivity on BTMRI is driven by category-level semantic information and lesion existence cues, rather than by spatial location. On BUSI, missing location descriptors lead to a larger decline, with MedCLIP Dice falling from 82.43\% to 71.58\%, indicating that spatial location information is more important for ultrasound images. These results show that text sensitivity depends on both the dataset and the specific semantic component. The text components that drive sensitivity in each modality require further investigation.


Fig.~\ref{fig:dice_box} further visualizes the distribution of Dice scores across the four datasets under different text perturbations. The box plots show that on ISIC and Kvasir-SEG, the Dice values of all models remain relatively stable across text conditions, indicating weak dependence on text guidance. In contrast, on BUSI and BTMRI, performance fluctuates dramatically when text is perturbed, confirming that these datasets exhibit much higher sensitivity to textual input. This stark contrast highlights that the degree of text sensitivity varies considerably across different medical imaging modalities and lesion characteristics, aligning with the quantitative findings in Table~\ref{tab:text_perturb}.
\section{Conclusion}

In this paper, we investigated the role of text in multimodal medical image segmentation. Systematic experiments across five VLMs and four datasets revealed that segmentation performance is largely insensitive to fusion module choice, suggesting visual representations dominate current multimodal models. We proposed an Evidence Decoupling Decoder (EDD) that decomposes image and text-modulated evidence while maintaining competitive performance. Perturbation experiments showed that text sensitivity is strongly dataset dependent. On BTMRI and BUSI, removing text caused severe degradation, while on ISIC and Kvasir-SEG text exerted relatively marginal influence. Notably, deleting location descriptors barely affected BTMRI performance but removing all text caused catastrophic failure, indicating category-level semantics rather than spatial cues primarily drive sensitivity on this dataset. We emphasize that perturbation results reflect model sensitivity rather than intrinsic text contribution, and that EDD characterizes learned representations rather than ground-truth attribution. Future work will pursue more rigorous modality attribution methods and adaptive multimodal frameworks.

%
%
%
%

\bibliographystyle{splncs04}
\bibliography{main}

\end{document}